\documentclass[conference]{IEEEtran}
\usepackage{times}

\usepackage[numbers]{natbib}
\usepackage{multicol}
\usepackage[bookmarks=true]{hyperref}

\usepackage[utf8]{inputenc} % allow utf-8 input
\usepackage[T1]{fontenc}    % use 8-bit T1 fonts
\usepackage{hyperref}       % hyperlinks
\usepackage{url}            % simple URL typesetting
\usepackage{booktabs}       % professional-quality tables
\usepackage{amsfonts}       % blackboard math symbols
\usepackage{nicefrac}       % compact symbols for 1/2, etc.
\usepackage{microtype}      % microtypography
\usepackage{bm}

\usepackage{times}
\usepackage{amsfonts} 
\usepackage{graphicx}
\usepackage{tabularx}
\usepackage{comment}
\usepackage{wrapfig}
\usepackage{amsmath}
\usepackage{amssymb}
\usepackage{array}
\usepackage{wrapfig,lipsum,booktabs}
\usepackage{hyperref}
\usepackage{booktabs}
\usepackage{multirow}
\usepackage{subfig}
\usepackage{setspace}
\usepackage{xcolor}
\usepackage{svg}

\begin{document}

\title{Keypoint Action Tokens Enable\\In-Context Imitation Learning in Robotics}

\author{\authorblockN{Norman Di Palo}
\authorblockA{The Robot Learning Lab\\
Imperial College London\\
London, UK\\
\texttt{n.di-palo20@imperial.ac.uk}}
\and
\authorblockN{Edward Johns}
\authorblockA{The Robot Learning Lab\\
Imperial College London\\
London, UK
}}

% avoiding spaces at the end of the author lines is not a problem with
% conference papers because we don't use \thanks or \IEEEmembership

%Homer Simpson\authorrefmark{2},
%Montgomery Scott\authorrefmark{3} and
%Eldon Tyrell\authorrefmark{4}}
%Telephone: (800) 555--1212, Fax: (888) 555--1212}
%\authorblockA{\authorrefmark{4}Tyrell Inc., 123 Replicant Street, Los Angeles, California 90210--4321}}

\twocolumn[{
\renewcommand\twocolumn[1][]{#1}
\maketitle
\begin{center}
    \centering
    \captionsetup{type=figure}
    \def\svgwidth{\textwidth}
    \vspace{-10pt}
  \includegraphics[width=1.\linewidth]{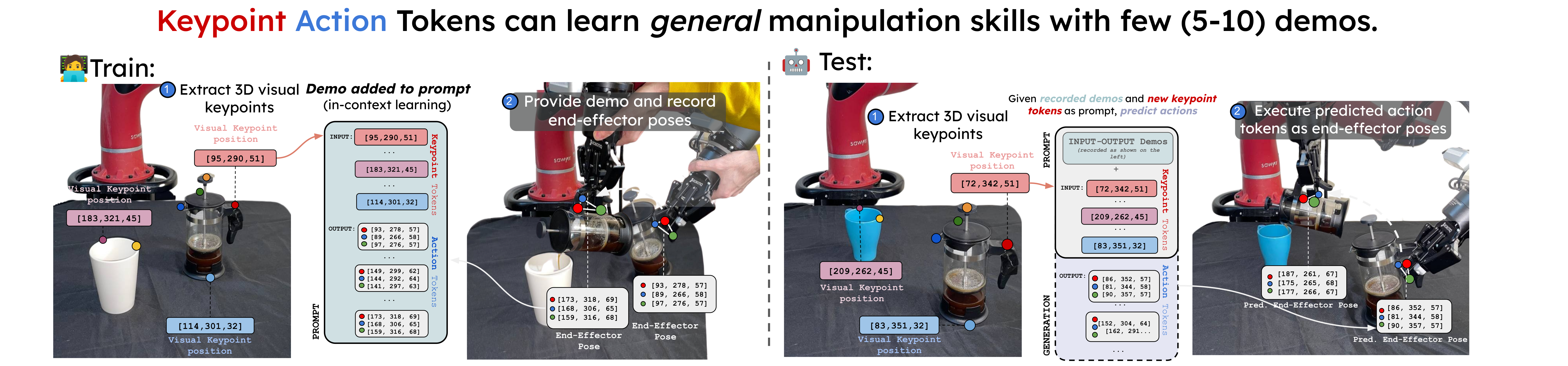}
    \caption{\footnotesize We introduce Keypoint Action Tokens (KAT), a method to repurpose off-the-shelf, large text-pretrained Transformers for efficient imitation learning, by representing sequences of observations and actions as sequences of keypoint-based tokens, converted into text and added to a prompt. KAT can learn general behaviours with as few as 10 demonstrations, and since no training is needed, learned skills can be deployed immediately after the demonstrations.}
    \label{fig:fig1}
\end{center}
}]

\begin{comment}
\begin{figure}[t]
    \centering
    \includegraphics[width=.5\textwidth]{figures/kp_fig1_2.png}
    \caption{}
    \label{fig:key-idea}
     \label{fig:retrieval}
\end{figure}
\end{comment}

%We demonstrate that it is possible to repurpose off-the-shelf Transformers pretrained on text to perform few-shot in-context imitation learning from human demonstrations, mapping visual inputs to sequences of low-level actions emulating the expert's behaviour.
\begin{abstract}
We show that off-the-shelf text-based Transformers, with no additional training, can perform few-shot in-context visual imitation learning, mapping visual observations to action sequences that emulate the demonstrator's behaviour. We achieve this by transforming visual observations (inputs) and trajectories of actions (outputs) into sequences of tokens that a text-pretrained Transformer (GPT-4 Turbo) can ingest and generate, via a framework we call Keypoint Action Tokens (KAT). Despite being trained only on language, we show that these Transformers excel at translating tokenised \textit{visual keypoint observations} into \textit{action trajectories}, performing on par or better than state-of-the-art imitation learning (diffusion policies) in the low-data regime on a suite of real-world, everyday tasks. Rather than operating in the language domain as is typical, KAT leverages text-based Transformers to operate in the vision and action domains to learn general patterns in demonstration data for highly efficient imitation learning, indicating promising new avenues for repurposing natural language models for embodied tasks. Videos are available at \href{https://www.robot-learning.uk/keypoint-action-tokens}{https://www.robot-learning.uk/keypoint-action-tokens}

\end{abstract}

\IEEEpeerreviewmaketitle

\begin{figure*}[t!]
  \centering
    \def\svgwidth{\textwidth}
  \includegraphics[width=0.99\linewidth]{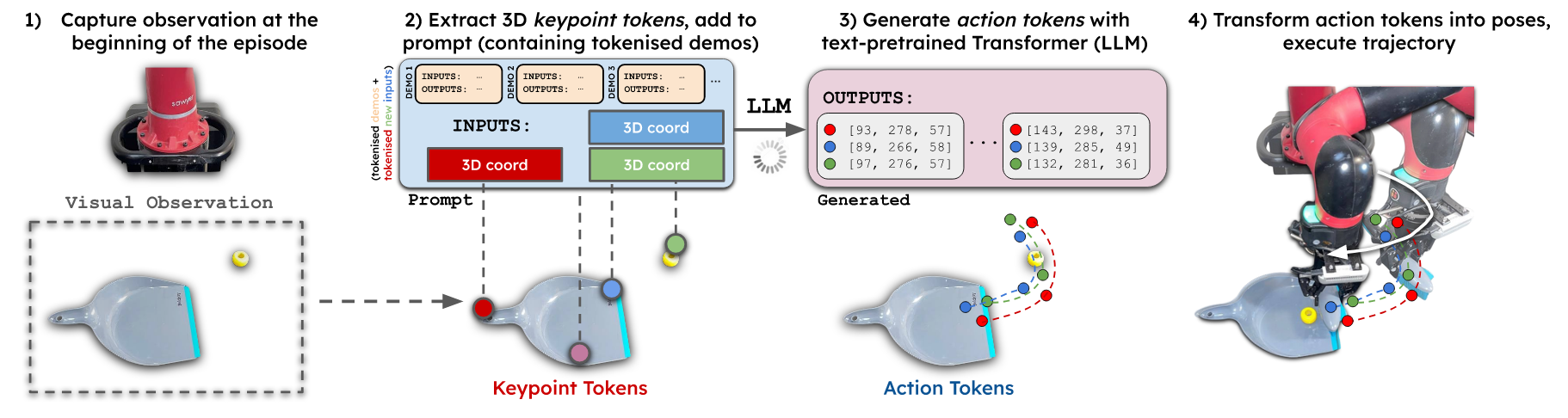}
  \caption{An illustration of our pipeline. \textbf{KAT} transforms a visual observation into a \textit{sequence of keypoint tokens}. From it, the text-pretrained Transformer (LLM) predicts \textit{action tokens}, that are then executed as a trajectory of poses.}
      \label{fig:pipeline}

\end{figure*}

\section{Introduction}

A robot's ability to learn new skills from just a few demonstrations, known as few-shot imitation learning, is important for scalable robot learning \cite{duan2017oneshot}. However, learning robust skills that can generalise to conditions not seen during the demonstrations is very challenging \cite{embodimentcollaboration2023open}. Whilst recent successes in imitation learning have shown that scaling up behavioural cloning can sometimes achieve such generalisation \cite{brohan2023rt2, embodimentcollaboration2023open}, this requires collecting huge datasets of demonstrations covering diverse objects and conditions, and we are still lacking effective methods for learning of entirely novel skills from just a few demonstrations.

Meanwhile, recent works have demonstrated how Transformers \cite{vaswani2023attention} excel in the ability to learn sequence-to-sequence patterns from
few examples \cite{brown2020language, olsson2022context}. Interestingly, a remarkable recent observation is that Transformers can effectively learn to emulate patterns
even when pre-trained on data from vastly different domains, acting therefore as \textit{general pattern machines}. In particular, large Transformers pre-trained on text have demonstrated the ability to quickly learn sequence-to-sequence patterns that go well beyond text-based reasoning tasks \cite{mirchandani2023large}. Building upon this observation, in this paper we now investigate the following: can we repurpose off-the-shelf large Transformers, pre-trained on language where data
is abundant, to act as \textit{few-shot imitation learning machines} in robotics, where data is scarce?

We study this by modelling imitation learning as a sequence-to-sequence prediction problem, with the input sequence being observations and actions from the demonstrations, plus the robot's current observation, and the output being the sequence of actions the robot should execute. However, a key challenge here is grounding text-based Transformers in vision, such that these sequences, which represent visual observations and end-effector actions, can be expressed as language tokens. We therefore introduce \textbf{Keypoint Action Tokens (KAT)} \ref{fig:fig1}, a framework that makes use of off-the-shelf Vision Transformers \cite{dosovitskiy2021vit} and text-based large Transformers \cite{brown2020language} to create a \textit{low-level language of robotics actions}. By converting the sequences of observations and actions into a \textit{sequence of tokens}, we show that a large Transformer, whilst pre-trained only on text, can learn to emulate complex physical behaviours from only a few example demonstrations ($\leq 10)$. 

We investigate Keypoint Action Tokens through a series of real-world experiments, providing insights into performance across a variety of few-shot learning challenges, such as spatial generalisation, generalisation to novel objects, learning multi-modal behaviours, and learning to perform trajectories with 6-DoF action spaces. We show that with Keypoint Action Tokens, off-the-shelf text-pretrained Transformers are able to perform few-shot learning on everyday tasks and are on par or superior to the current state-of-the-art in imitation learning (diffusion policies \cite{chi2023diffusion}). To the best of our knowledge, this is the first work to propose a comprehensive method to perform imitation learning from vision using an off-the-shelf large text-pretrained Transformer as the core imitation learning algorithm, with no additional training. A key contribution of our work is strong evidence that the \textbf{progressive evolution of capabilities of large pretrained Transformers is leading to the emergence of more and more general and efficient pattern learning machines, that can directly be used off-the-shelf to tackle sequence-to-sequence imitation learning \textit{without the need for training on any robotics data}}. The ability to repurpose large networks trained on language domains, where data is abundant, is a promising avenue to unlock unprecedented learning efficiency in robotics, where data is scarce.

\section{Related Work}

%general LLMs citations
%transformers can learn patterns

\textbf{Few-shot Imitation Learning.} Few-shot imitation learning is one of the main desiderata in robot learning, and more generally, in machine learning. Behaviour Cloning, where a neural network is trained over a dataset of observation-action pairs coming from an expert demonstrator, often requires hundreds of laborious demonstrations to learn tasks. Over the years, many methods have been proposed to tackle this problem. \cite{duan2017oneshot} proposed to train a network to explicitly take in a demonstration as input, and output actions that would follow the behaviour demonstrated. \cite{james2018task} proposed to learn a Siamese network on pairs of demonstrations of the same behaviour to learn task-specific embeddings for one- or few-shot learning. \cite{clavera2018modelbased} proposed to train several imperfect dynamics model to train a policy to perform efficient adaptation to real-world data. All these methods, however, still rely on gathering a substantial amount of interaction data. We propose to use the emergent pattern completion abilities of large Transformers pretrained solely on text, where no robotics data is used to pretrain or finetune it.

\textbf{Finetuning Large Pretrained Models} An effective paradigm that emerged from the recent research is to first pretrain large models on enormous, task-agnostic datasets, and then perform finetuning with substantially smaller datasets representing the task at hand. Such models are referred to as Foundation Models \cite{bommasani2022opportunities}. While being extremely successful in the fields of computer vision and natural language processing, the lack of vast robotics datasets hindered the creation of robotics specific Foundation Models. While \cite{brohan2023rt2, embodimentcollaboration2023open} demonstrated promising results by finetuning large Vision-Language Models on robotics data, learning new tasks is still inefficient and laborious. With Keypoint Action Tokens, we show that no fine-tuning is needed and thus we can achieve highly efficient in-context learning.

\begin{figure*}[t!]
    \centering
    \includegraphics[width=.99\textwidth]{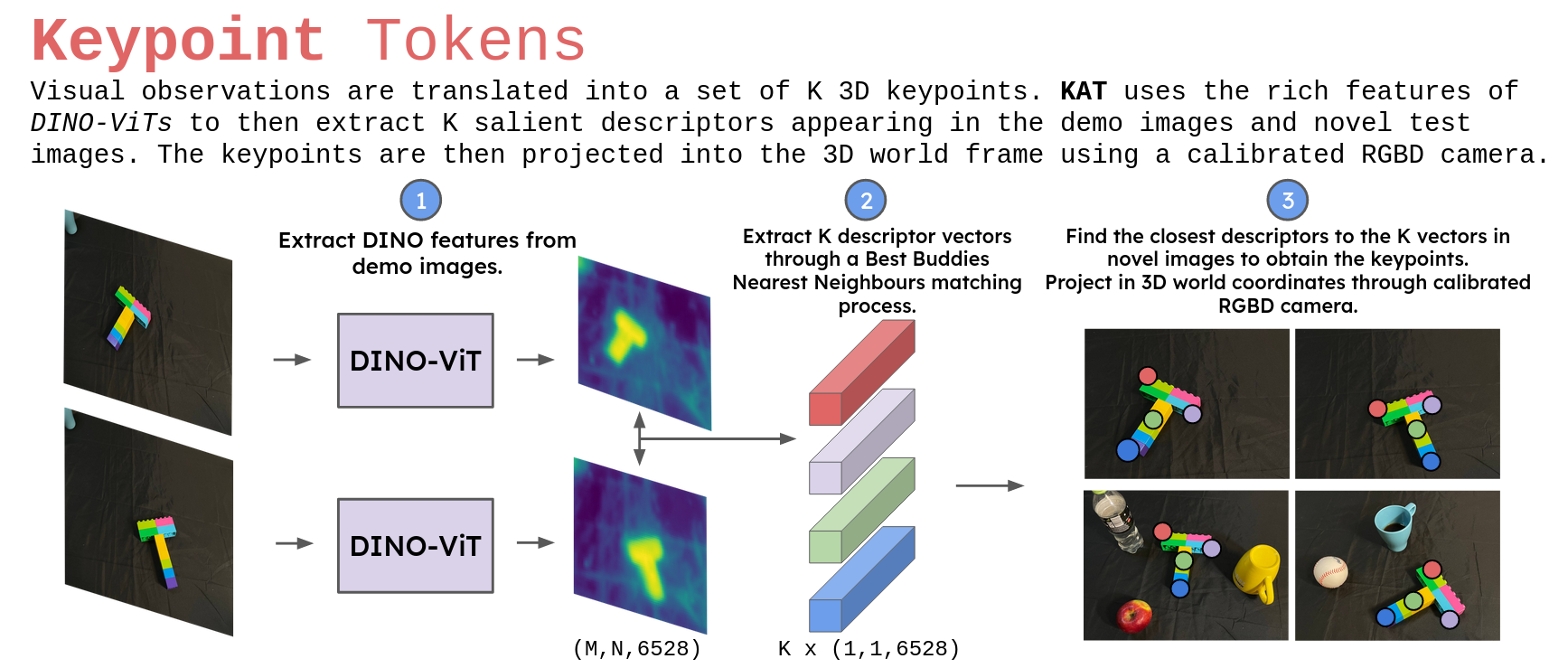}
    \caption{Illustration of the extraction pipeline of \textit{keypoint tokens}.}
    \label{fig:keypoint-tokens}
\end{figure*}

\textbf{Use of Large Language Models in robotics as planners.} The exponential progress in capabilities of text-pretrained Transformers, or Large Language Models (LLMs), to interpret and reason over text generated a plethora of robotics works leveraging such models. \cite{ahn2022saycan, liang2023code, vemprala2023chatgpt, huang2022language, driess2023palme} use LLMs as high-level planners, obtaining a list of steps to solve tasks in language form. \cite{huang2023voxposer, yu2023language} leverage LLMs to map a high-level description of a task in natural language into rewards that can be used by a trajectory optimiser. While effective, these methods rely on LLMs to generate high-level plans or guidance for external optimisers, and therefore need additional techniques to ground these outputs into actions. In our work, we \textit{do not use LLMs for natural language}, but instead exploit their pattern learning capabilities to directly process sequences of observations and generate executable low-level trajectories. We therefore shed light on a counter-intuitive phenomenon:\textit{ \textbf{one of the most effective uses of LLMs in robot learning is achieved \textit{without the use of natural language, neither in the inputs nor in the outputs}.}}

\textbf{Language Models that output robotic actions.} \cite{brohan2023rt2} finetunes a Vision-Language Model with robotics data in the form of language-annotated demonstrations. They therefore obtain a model that can receive as input a language instruction, and can output, similarly to our method, actions that can directly be executed by the robot. The main differences, however are that 1) we do not need finetuning, instead employing off-the-shelf models, and 2) while their model receives language instructions, our method can perform \textit{in-context learning} on low-level demonstrations, therefore acting as a general imitation learning machine.  \cite{kwon2023language} demonstrated the ability of LLMs to directly output trajectories of executable actions instead of high-level textual plans. However, this can only generate and execute trajectories based on language instructions and thus being unable to learn from demonstrations.
\cite{mirchandani2023large} was one of the first works to show the general pattern learning abilities of LLMs. They demonstrated the capability of LLMs to autoregressively complete a partial trajectory received as an input, therefore learning patterns beyond human language. However, in contrast to their work, we propose a full imitation learning framework that receive complex visual inputs and replicate expert behavior on many everyday tasks, that present a set of unique challenges we will later discuss. To the best of our knowledge, we are the first work to propose a complete framework that can learn from demonstrations in the forms of sequences of visual observations and actions using off-the-shelf text-pretrained Transformers.

\textbf{In-Context Learning} Scaling the training of large Transformers, especially on language data, led to the emergence of in-context learning \cite{brown2020language}. While the explanation for the phenomenon is not entirely clear, recent works have investigated its emergence in simple settings \cite{olsson2022context}. \cite{vonoswald2023transformers} investigated the parallels between in-context learning, a purely feed-forward adaptation mechanism with no weights updates, and gradient descent, hypothesising that Transformers learn to implicitly apply a gradient update via attention heads. These observations inspired this work, and the use of Transformers and their in-context learning ability as general imitation learning machines.
Note that Transformers are not the only way to obtain in-context learning. \cite{vosylius2023fewshot} trained a graph neural network to learn object alignments when conditioned on a few demonstrations in an in-context learning manner. However, they represent imitation learning as learning optimal objects alignments, while we represent policies exclusively as a sequence of robot poses, proposing both a new input-output representation (keypoint and action tokens) and the use of pretrained Language Models as in-context imitation learning machines.

%However, in recent years, a paradigm have achieved the most success: pretraining large networks on enormous task-agnostic datasets, and then finetuning these models on a smaller dataset of data coming from the task at hand. These models are referred to as Foundation Models. Despite their great success in the fields of computer vision and natural language processing, such models have not yet been developed in the field of robotics due to the relative scaricy

%LLMs in robotics

%In context learning in robotics

%The scarcity of robotics data led the community to repurpose foundation models. Vision Models are often used as backbone feature extractors before training a BC policy. LLMs are often used as high level planner, or as language conditioned trajectory generators. We show that there is a more performant way of using both, that does not require any language input but is instead grounded and conditioned on low level actions.

%RT-2 finetuned a VLM on actions, creating a VLA and effectively obtaining a "language of actions". However, 1) it is computationally expensive and not open source 2) they obtain a zero-shot language conditioned policy, while we propose a more versatile and plastic in-context imitation learning method that can learn tasks that may difficult to describe.

%In context learning has been tried with different modalities/results. Mention IGA by Vitalis, but it requires an adhoc network training, plus it frames IL as poses matching, while we do actions, therefore being more general.

\section{Method}
We formulate few-shot imitation learning as a sequence-to-sequence problem. Let $\mathcal{O}$ represent the sequence of observations received by the robot and $\mathcal{A}$ the sequence of end-effector actions - in our case a trajectory of $SE(3)$ poses $T^\mathcal{W}$ expressed in the world frame $\mathcal{W}$. Let $\mathcal{D} = [d_1, \dots, d_n]$ be a sequence of demonstrations given by a human expert, with each $d_i = [\mathcal{O}_i, \mathcal{A}_i]$ representing the desired sequence of actions given a certain sequence of observations. Our goal is to find an effective \textit{imitation learning machine} $\mathcal{M}(\mathcal{D}, \mathcal{O}_t) \rightarrow \mathcal{A}_t$, with $\mathcal{O}_t$ being a novel sequence of inputs and $\mathcal{A}_t$ being a predicted sequence of actions that would effectively emulate the expert behaviour implicitly represented in $\mathcal{D}$.

In this work, the robot records a visual observation, $o_i$, at the beginning of the episode. Therefore, for each demonstration $d_i$, the robot collects an observation $o_i$ and a sequence of actions $a_{i,1:T}$. When deployed, the robot's task is to record a new observation $o_j$ and compute a sequence of actions $a_{j,1:T}$ that emulate the expert behaviour observed in $\mathcal{D}$. The pipeline is illustrated in Fig. \ref{fig:pipeline}. Recording and using a single visual observation is a design choice of this work and not an inherent limitation of our method, which could receive a new visual observations every $T$ steps, and compute a new trajectory of actions conditioned each time.

The current literature often decomposes the problem in two steps: first, the output of $\mathcal{M}(\mathcal{D})$ is a policy network $\pi_\theta$ trained on $\mathcal{D}$; second, the sequence of actions is computed as $\pi_\theta(\mathcal{O}_t) \rightarrow \mathcal{A}_t$. We demonstrate, however, that this decomposition is not necessary: we can instead leverage the \textit{in-context learning} abilities of pretrained Transformers to input both $\mathcal{D}$ and $\mathcal{O}_t$ to the network, and then compute $\mathcal{A}_t$ emulating the patterns observed in $\mathcal{D}$.

More specifically, \textit{in-context learning} \cite{olsson2022context, brown2020language, vonoswald2023transformers} commonly refers to the ability of certain networks to perform few-shot learning \textit{without} the need to update their parameters. The network is conditioned on both a few example datapoints, and the new input from which it needs to compute an output following the behaviour expressed in the examples, taking the form $f_\phi(\mathcal{O}_t, \mathcal{D}) \rightarrow \mathcal{A}_t$. While not entirely exclusive to them \cite{vosylius2023fewshot}, the recent literature suggests that this ability emerges from large Transformers trained to perform autoregressive pattern completion on vast datasets, the most successful case being language \cite{olsson2022context, brown2020language}.

The core finding of this work is that pretraining or finetuning using robotics data \textit{is not needed}, and we can instead \textbf{repurpose a large Transformer pretrained entirely on textual data}, (a Large Language Model), without any additional finetuning, and it can act as an extremely efficient \textit{imitation learning machine} $\mathcal{M}$.

To unlock these abilities, we first need to transform $\mathcal{O}$ and $\mathcal{A}$ into a data format that can be interpreted by text-pretrained Transformers, that ingest \textit{text}.
In this work, $tokenising$ refers to transforming observations and actions into a form that can be given as input to a text-pretrained Transformer/Large Language Model, as we will shortly describe in more detail.

%A \textit{token} is a small sequence of characters, represented as an integer via a lookup table. Each integer represents a unique sequence of characters. Intuitively, each word can be seen as composed by one or more tokens. With \textit{tokenisation} we refer to the process of transforming sequence of characters into the corresponding sequence of integers. 

Our goal is to represent $\mathcal{O}$ and $\mathcal{A}$ as a sequence of tokens, finding a \textit{language} that represents observations and actions effectively. To do so, we introduce \textbf{Keypoint Action Tokens (KAT)}. Our framework transforms:
\begin{itemize}

\item visual observations into a \textit{sequence of 3D visual keypoints}, which are then tokenised as strings of characters: we define those as \textbf{keypoint tokens} (Fig. \ref{fig:keypoint-tokens}). 

\item end-effector trajectories of $SE(3)$ poses, into a \textit{sequence of 3D points} fixed in the end-effector frame, which are also tokenised into a sequence of characters: we refer to those as \textbf{action tokens} (Fig. \ref{fig:action-tokens}).
\end{itemize}

 We will now describe both processes in more detail.  

\subsection{Keypoint Tokens}

The observations received by the robot are in the form of top-down RGBD images $o_i$ of dimension $(512,512,4)$, taken from a calibrated RGBD camera at the beginning of each episode. To accelerate the learning process, we aim to extract lower-dimensional data from the sequence of pixels. In particular, we extract from each image $o_i$ a sequence of \textit{visual keypoints} $k_{1:K}$, as illustrated in Fig. \ref{fig:keypoint-tokens}. To do so, we leverage recent Vision Foundation Models: large vision networks pretrained on vast datasets of images. We use DINO Vision Transformers (DINO-ViTs) \cite{caron2021dino}, a family of networks that has been shown to be able to extract information-rich descriptors from images, representing both the semantic and geometrical information present in the image. We follow the method proposed in \cite{amir2022dinokp}: we first compute the dense descriptor tensors $E$ of shape $(H,W,6528)$ (where $H,W < 512$ and depend on the patch size selected) from each image collected during the demonstrations in $\mathcal{D}$, by inputting each image into the DINO-ViT. We then extract, through a Best Buddies Nearest Neighbours method \cite{amir2022dinokp}, a sequence of $K$ descriptors, e.g. one-dimensional vectors of shape $(1,1,6528)$ corresponding to a particular patch in one of the demonstration images among the $H \times W$ patches and corresponding descriptors. Intuitively, the extracted descriptors represent parts of the image that are common to all the collected observations, therefore encoding a particular semantic or geometrical structure (\textit{e.g. the handle of a mug, the cap of a bottle}). This process is entirely autonomous. Once we have obtained these $K$ descriptors, we can extract the visual keypoints from test time observations $o_t$. To do so, we extract the dense descriptor tensor $E_t$ from $o_t$, and find the closest descriptor to each of the $K$ stored descriptors via nearest neighbour search, obtaining $K$ 2D spatial locations $h_k, w_k$. These represent 2D keypoints in the image. Through the calibrated RGBD camera, we can first obtain the depth component of the 2D pixels, making them 3D, and then transform them into world frame $\mathcal{W}$ coordinates from camera coordinates. The final output of this process is a \textit{sequence of 3D visual keypoints in world coordinates}. By representing these numbers as characters we obtain the \textbf{keypoint tokens $k$}, which will then be used as the input sequence $\mathcal{O}$.

\subsection{Action Tokens}
\label{sec:action-tokens}
\begin{figure}[t]
    \centering
    \includegraphics[width=.5\textwidth]{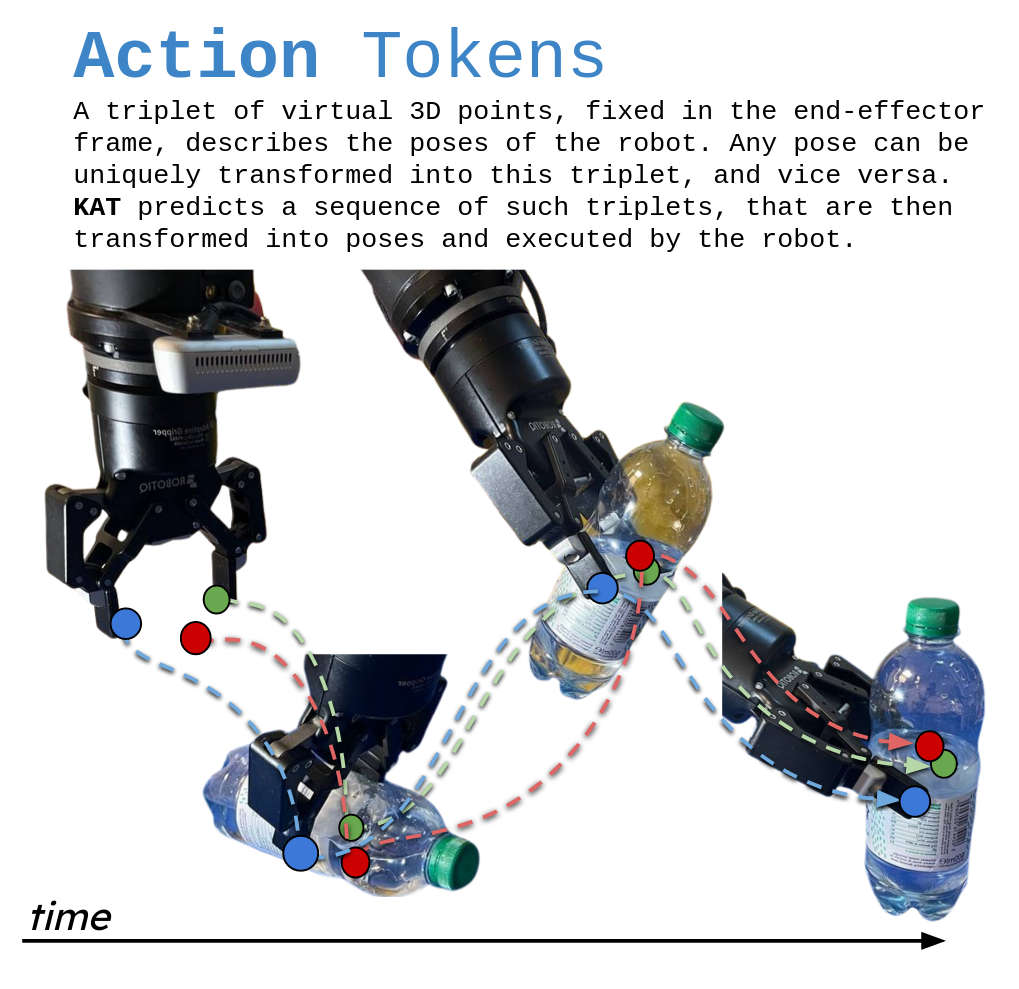}
    \caption{Illustration of the \textit{action tokens}, used in this work to represent $SE(3)$ end-effector poses.}
    \label{fig:action-tokens}
    \vspace{-10pt}
\end{figure}

Given input $\mathcal{O}$, the human experts provide a sequence of actions $\mathcal{A}$, that is a trajectory of end-effector poses, which demonstrates a desired behaviour. In our work, we represent end-effector poses as a triplet of 3D virtual points expressed in meters $\bm{e}_1 = [0.05,0.0,0.0], \bm{e}_2 = [-0.05,0.0,0.0], \bm{e}_3 = [0.0,0.1,0.0]$, represented collectively as $\bm{e} = [\bm{e}_1^T, \bm{e}_2^T, \bm{e}_3^T]$, defined in the end-effector frame $\mathcal{E}$. Each end-effector pose is then represented as the triplet of points in world frame $\bm{e}^\mathcal{W} = \mathcal{T}^\mathcal{W}_\mathcal{E} \bm{e}$, where $\mathcal{T}^\mathcal{W}_\mathcal{E}$ is obtained through the robot's forward kinematics. The mapping is unique and invertible: each pose can be uniquely represented as a triplet $\bm{e}^\mathcal{W}$, and vice versa, from each triplet we can recover the corresponding pose. More details are provided in the Supplementary Material. 

Note that representing an $SE(3)$ pose as a triplet of 3D points is mathematically  \textit{redundant}. Why did we opt for this redundancy? The reason is that, in this way, each 3D point lies in the same Euclidean space of the 3D keypoints tokens, therefore assisting with pattern recognition. On the other hand, the use of rotational representations, like angle-axis, Euler angles or quaternions, while requiring fewer numbers, would have required to implicitly learn a mapping from a Euclidean space to a different manifold. In our Experiments section we will empirically prove these benefits.

In Fig. \ref{fig:action-tokens} we illustrate a trajectory of the end-effector placing a bottle upright, showing the triplet of points moving in space. Each triplet $\bm{e}^\mathcal{W}$ is represented by 9 numbers, which we can represent as characters to obtain \textbf{action tokens $a$}. A trajectory $\mathcal{A}$ is therefore represented as a sequence of action tokens.

While the \textit{action tokens} computed from the demonstrations always form a triangle with a predefined shape, these tokens are independently predicted at test time and so this geometrical constraint may not always hold. We therefore find the closest triangle via a Least Squared Error approach, and from that compute the end-effector pose.

To represent the gripper status, we add an additional number $\delta_g$ to the 9 representing the pose, indicating an open gripper if $\delta_g = 0$ and a closed gripper if $\delta_g = 1$. At test time, we round possibly noisy predictions as $\delta_g \ge 0.5 \rightarrow \delta_g = 1$, and $\delta_g < 0.5 \rightarrow \delta_g = 0$. This gripper representation is used in the same way in all the baselines we will later introduce.   

This representation is similar to the one proposed in \cite{khansari2020action}, studying efficient representations for grasping, that however operates in pixel-space: they in-paint a triplet of pixels on the visual observation representing where the end-effector would move. They then sample possible pixel-locations of these triplets, and use a learned network to verify if that would lead to a successful grasp. Our representation is instead in 3D world-frame Euclidean coordinates, and we output a sequence of such poses to perform tasks well beyond grasping.  

\subsection{In-Context Imitation Learning via Pretrained Transformers}
Transformers trained on textual datasets to perform language modelling, referred to as Large Language Models, receive a \textit{prompt} $p_i$ as an input, that is a sequence of tokens, and autoregressively compute a new sequence of tokens $t_i \leftarrow f(p_i)$ that maximise the likelihood $p(t_i | p_i)$. While originally trained to complete textual prompt, and therefore to produce coherent new text given a prompt emulating the natural language probability distribution appearing in the data, these Transformers were observed to gain the emergent capability to perform general pattern completion well beyond the domain of natural language \cite{mirchandani2023large}. 
In each training episode the robot collects a visual observation and a trajectory of actions recorded by the expert demonstrator: the observation is transformed into a  \textit{sequence of keypoints tokens} and the actions are transformed into our chosen triplet representation as a \textit{sequence of action tokens}. 
At test time, the robot collects an observation that is also transformed into a new \textit{sequence of tokens}. When receiving these sequences of keypoint and action tokens as a prompt $p_i$, the Transformer therefore will be conditioned to output, given the novel sequence of keypoint tokens, a new sequence of action tokens that mimics the behaviour demonstrated by the expert in $\mathcal{D}$. The action tokens, i.e. the triplets $\bm{e}^W$, will then be transformed into $SE(3)$ poses of the end-effector and executed sequentially.

%In the previous sections we described how we can transform both sequences of inputs $\mathcal{O}$ and sequences of outputs $\mathcal{A}$ into a representation that can be fed into a text-pretrained Transformer, using respectively sequences of \textbf{keypoints tokens} and sequences of \textbf{action tokens}. 

As such, in this work we tackle the sequence-to-sequence few-shot imitation learning problem by first transforming all sequences in $\mathcal{D}$ into keypoint action tokens $[k_{d}, a_{d}]_{d=1}^D$, appending each one to the list of tokens $p_t$ given as input to the Transformer. We also represent the new observation $o_t$ for the task at hand as a sequence of keypoint tokens $k_{t}$, and append those to $p_t$ as well. The robot will then execute the sequence of predicted action tokens $a_t \leftarrow f(p_t)$ with $p_t = [[k_{d}, a_{d}]_{d=1}^D, k_t]$, transformed into a trajectory of $SE(3)$ poses.

The advantages of using a large pretrained Transformer as an \textit{imitation learning machine} are twofold:
\begin{itemize}
    \item \textbf{Data efficiency}: the vast pretraining allows the Transformer to \textbf{more efficiently extract patterns from sequences} than training policy networks on that data from scratch.
    \item \textbf{Time efficiency}: \textit{in-context learning} only requires a single forward pass through the network and no weight updates. This results in the ability to \textbf{deploy the robot immediately after receiving the demonstrations}, in contrast to the need to finetune networks on the new data.
\end{itemize}

\begin{figure*}[t!]
    \centering
    \includegraphics[width=.99\textwidth]{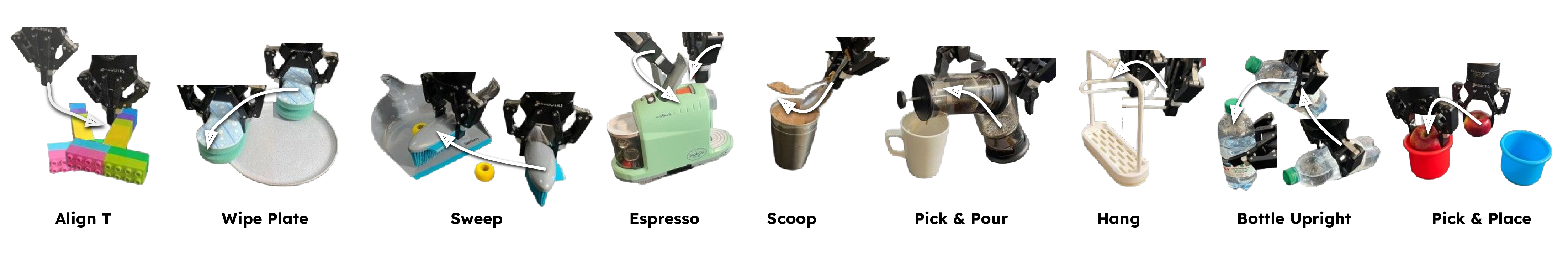}
    \caption{The tasks we evaluated our method and the baselines on.}
    \label{fig:tasks}
\end{figure*}

\section{Experiments}

Our paper proposes the use of text-pretrained Transformers as few-shot imitation learning machines with our Keypoint Action Tokens framework. The core investigation performed in our experiments is therefore aimed at answering two main questions: \textbf{1)} How does \textbf{KAT} compares against current state-of-the-art imitation learning frameworks? \textbf{2)} What is the optimal design of keypoint and action tokens in order to maximise performance, and why?

Additionally, we also provide a series of ablation studies and analyses designed to shed light on more properties of our method, like robustness to distractors. Additional details can be found in the Supplementary Material, including videos.

\subsection{Experimental Setup}
We run our experiments using a Sawyer robot, interacting with objects on a table in front of it. The end-effector is a Robotiq 2F-85, on which we mount an Intel Realsense D435 RGBD camera.
In every interaction episode, the robot records one visual observation at the beginning of the task, then computes a trajectory of actions, and finally executes those actions. The end-effector is moved 90cm above the table at the beginning of each episode to record an image observation.

During demo collection, the robot first captures an observation, and then the human expert manoeuvres the end-effector through kinesthetic teaching, while its $SE(3) $ poses are recorded. Each demo therefore collects an RGBD observation and a sequence of end-effector poses. When collecting demos, the positions of the objects for each task are randomised from within the starting state distribution of the task. More details can be found in the Supplementary Material.

At test time, the objects are positioned on the table in novel configurations, designed to test both \textit{interpolation} and \textit{extrapolation} abilities with respect to object poses during training. The robot captures an observation as above, and then computes a series of end-effector poses.

We use \textbf{GPT-4 Turbo} \cite{openai2023gpt4} as the Large Language Model, but later compare its performance to other models. In the Supplementary Material we provide more details about the way we give inputs to it and the way we process its outputs.

\subsection{Tasks}

\begin{table*}
    \centering
    \scalebox{0.85}{%
    \begin{tabular}{|c|c|c|c|c|c|c|c|c|c|c|} \hline 
         \textbf{Method/Task}&  \textbf{Align T} &  \textbf{Wipe Plate}&  \textbf{Sweep}&  \textbf{Espresso}&  \textbf{Scoop}&  \textbf{Pick \& Pour}&  \textbf{Hang}&  \textbf{Bottle}& \textbf{Pick \& Place} & \textbf{Avg.}\\ \hline 
         \textbf{KAT (Ours)}&  \textbf{0.8}&  \textbf{0.8}&  0.7&  \textbf{0.4}&  \textbf{0.6}&  \textbf{0.7}& \textbf{0.7}&  \textbf{0.6}& \textbf{0.8} & \textbf{0.68}\\ \hline 
         \textbf{Diff. Policies}&  0.0&  0.2&  0.1&  0.0&  0.0&  0.0&  0.2&  0.1& 0.3 & 0.1\\ \hline 
         \textbf{KeyAct-DP}&  0.4&  0.7&  \textbf{0.8}&  \textbf{0.4}&  0.4&  \textbf{0.7}&  \textbf{0.7}&  0.4& \textbf{0.8}& 0.59\\ \hline
         \textbf{KeyAct-MLP}&  0.3&  0.4&  0.7&  0.3&  0.4&  0.6&  \textbf{0.7}&  0.4& 0.7 & 0.5\\ \hline
    \end{tabular}}
    \caption{Success rates with 10 demonstrations for each task, for KAT and the baselines (best in bold).}
    \label{tab:main_results}
\end{table*}

We thoroughly evaluate our method by choosing a family of everyday manipulation tasks and measuring its ability to efficiently replicate the expert behaviour, generalising after just a few demonstrations. The criteria we adopted to choose the tasks are: \textbf{1)} to replicate tasks that appeared in the recent literature, \textbf{2)} to measure \textbf{KAT}'s ability to tackle a series of challenges that appear in everyday tasks. In particular, we chose tasks that collectively measure \textbf{generalisation to novel shapes, multi-modality, precision, dexterity, multi-stage execution, and 6D trajectories}. The tasks are the following:
\begin{itemize}
    \item \textbf{Align T}: Inspired by \cite{chi2023diffusion}, the robot needs to move a T-shaped object to align it to the axes of the table. The task requires \textit{non-prehensile} abilities and the ability to model \textit{multi-modal} distributions: demos are provided following different strategies (pushing the vertical or the horizontal part to align the T), and the robot needs to commit to a single one at test time.
    \item \textbf{Wiping a Plate}: The robot needs to follow the edge of a plate with a sponge in a circular motion. This task requires \textit{multi-modal} reasoning: half the demos are given in clockwise direction, half are given counter-clockwise. The robot needs to commit to one at test time. We test for both interpolation and extrapolation abilities by also using unseen plates at test time of unseen dimensions.
    \item \textbf{Sweep}: The robot needs to sweep an object into a dustpan. The relative positions of dustpan and object are randomised, therefore the robot needs to compute an effective trajectory to bring the object into the dustpan. We test for generalisation by also using a novel dustpan and objects to sweep at test time.
    \item \textbf{Espresso}: The robot needs to insert an espresso capsule into a toy espresso machine, and then close its lid. This task requires noteworthy \textit{precision}, and is composed of \textit{multiple stages}.
    \item \textbf{Scooping}: The robot needs to scoop some chocolate powder from a cup with a spoon. The task was proposed in \cite{zhou2023toto} as a challenging, dexterous task. We test for generalisation by using also unseen cups at test time.
    \item \textbf{Pick and Pour}: The robot needs to pick up a French Press and pour coffee into a cup. The pouring part of the task was proposed in \cite{zhou2023toto} as a \textit{challenging, dexterous task}. By also picking up the French Press, the task requires \textit{multiple stages}.  We test for generalisation by also using unseen cups at test time.
    \item \textbf{Hang}: The robot needs to hang a clothes hanger onto a horizontal support. The task requires to reach a \textit{precise position and height}, otherwise the hook would not hang on the support.
    \item \textbf{Put Bottle Upright}: The robot needs to pick up a horizontally placed bottle and place it upright on the table. This is also inspired from a task in \cite{chi2023diffusion}, where a mug has to be flipped. The task requires a \textit{non-trivial, 6D trajectory} to be completed, so that the bottle is stably placed. We test for generalisation by also using unseen bottles at test time.
    \item \textbf{Pick and Place}: The robot needs to pick up an apple and place it in a red bowl, with a blue bowl acting as a visual distractor. We test for generalisation at test time by both using an orange instead of an apple, and a purple bowl as a distractor instead of the blue bowl. This task requires the ability to \textit{ignore distractors} and understand which objects are useful, and which must be ignored. 
\end{itemize}

Additional details, like success criteria, are listed in the Supplementary Material. 

\subsection{Baselines}

As the goal of our work is to demonstrate the efficiency of Large Language Models (large text-pretrained Transformers) to effectively act as \textit{imitation learning machines} by learning to emulate expert behaviour after a few demonstrations, we compared our approach with Diffusion Policies \cite{chi2023diffusion}, a state-of-the-art general imitation learning algorithm from the recent literature, that was already demonstrated to surpass a series of recent techniques in terms of learning efficiency and generality.

We compare to two versions of this. First, the original image-based \textbf{Diffusion Policies} method proposed in \cite{chi2023diffusion}. Second, a modified version were we provide our proposed \textit{keypoints} as input and \textit{actions representation} as outputs (\textbf{KeyAct-DP}), therefore going from a $(512,512,4)$ input space to $K$ 3D visual keypoints, and from the original position and orientation representation of \cite{chi2023diffusion} to a triplet of 3D points uniquely representing an end-effector pose. We use the original authors' code, and optimise hyperparameters to maximise performance on our tasks.

Furthermore, we compare against using a feed-forward network to learn to map the same keypoints into a sequence of action tokens, (\textbf{KeyAct-MLP}).

These baselines allow us not only to compare \textbf{KAT} to the state-of-the-art in the field, but also to measure the contribution coming from the \textit{keypoints} and \textit{actions} representation we propose.

\subsection{Results on Few-Shot Imitation Learning}
\label{sec:few-shot-results}
\textbf{Can a text-pretrained Transformer/Large Language Model be used to perform in-context imitation learning of expert behaviours?} This is the core question we explored in this work. To answer this, we evaluated \textbf{KAT} and the baselines on the 9 tasks reported above. For each task, we provide \textbf{10 demonstrations and execute 10 test time episodes by randomising the position of the involved objects}. 5 episodes are tested with the same object(s) seen during training, while the remaining 5 are tested with \textit{unseen objects} where possible (as listed above, the only exceptions being \textit{Align T, Espresso} and \textit{Hang}). We extract $K=10$ \textit{keypoint tokens} from the observation recorded at the beginning of the episode, and record the demonstrations actions at $4$ Hz, therefore storing 4 poses per second. In later section, we will show the effect of these hyperparameters on the overall performance.

We show success rates in Table \ref{tab:main_results}. These results provide a series of interesting insights. First, they show that \textbf{KAT} \textbf{performs at the level of the state-of-the art in imitation learning, with better performance on certain tasks}. However, KAT does not need any training after receiving the demos, unlike the baselines that require $\sim$10 minutes of training per task. Second, our novel representation of observations and actions as \textit{3D points}, dramatically improves the performance of both Diffusion Policies over the original image-based Diffusion Policy method, and also make a simple MLP baseline perform better than end-to-end Diffusion Policies. Qualitatively, we also observed that Diffusion Policies struggled more to extrapolate to regions of the environment or object not covered during the training process.

\begin{figure}[t]
    \centering
    \includegraphics[width=.5\textwidth]{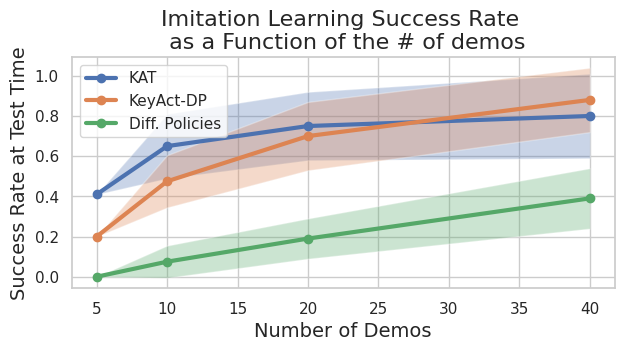}
    \caption{Success rate of each method as a function of the number of demos. While \textbf{KAT} outperforms the baselines in the few-shot regime ($\le 20$ demos), in-context learning struggles to improve as the number of demos increase even more. Plot shows mean and standard deviation across tasks.}
    \label{fig:num-demos}
\end{figure}

\textbf{How does the performance change as the number of demonstrations increases?}
In our previous experiments we provided 10 demonstrations for each task. To investigate the performance as a function of the number of demonstrations received, a key property in few-shot imitation learning, we chose a subset of the 9 tasks (\textit{Align T, Wipe Plate, Espresso, Bottle Upright}) and provided 5, 10, 20 and 40 demonstrations for each task. We then compared the performance of KAT and the baselines.
The results are plotted in Fig. \ref{fig:num-demos}. Diffusion Policy improves with the number of demonstrations, but at a very low rate and performance is poor even with 20 demonstrations, as the input space is substantially high-dimensional. Both KAT and KeyAct-DP achieve very good performance, with KAT's performance evolving more quickly than KeyAct-DP. However, when receiving 40 demonstrations, we can see that KeyAct-DP surpasses KAT on certain tasks. This suggests an important insight: pretrained Transformers can excel at in-context learning of sequence-to-sequence patterns few-shot, making them particularly useful for low-data regimes, but currently do not scale well as further in-context data is provided. We hypothesise that with this amount of data, in-context learning stops working optimally, and becomes a bottleneck. Instead, we believe that, given this amount of data, it may be more appropriate for Transformers to be finetuned to further improve performance, albeit the process would be more laborious that in-context learning. Nevertheless, finetuning LLMs on Keypoint Action Tokens data is an interesting direction for future work.

\begin{table}
    \centering
    \begin{tabular}{|c|c|c|c|c|} \hline 
          &  \textbf{Clean }& \textbf{Distractors}&  \textbf{Diff.  BG}& \textbf{Both}\\ \hline 
         \textbf{KAT (Ours)}& 0.65 & 0.6&  0.62& 0.58\\ \hline
    \end{tabular}
    \caption{Success rates for our method showing the effect of distractors and different background (BG).}
    \label{tab:distractors}
\end{table}
\textbf{Can our method handle visual distractors?}
To study robustness to distractors, we evaluate KAT on the \textit{Align T, Wipe Plate, Espresso, Bottle Upright} tasks, providing 10 demonstrations per task, but at test time we add distractor objects, change the background, or both. For each of these scenarios, we run 10 test time episodes, reporting the results in Table \ref{tab:distractors}. 
These results showcase the robustness of KAT to visual distractors. This ability is obtained as a combination of two factors: from a vision perspective, the descriptors extracted from the observations via DINO-ViTs are robust to perturbations and have high dissimilarity across different objects. From a sequence-to-sequence pattern learning perspective, the pretrained Transformer is able to infer the \textit{keypoints tokens} most relevant to the task by comparing the inputs-outputs relations in the demonstrations, therefore ignoring possible \textit{keypoint tokens} that are less robust to unrelated visual changes.   

\begin{figure}[t]
    \centering
    \includegraphics[width=.5\textwidth]{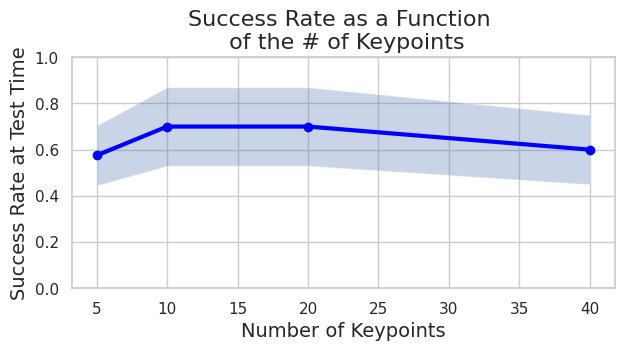}
    \caption{Success rate of our method as a function of the number of keypoints extracted. Plot shows mean and standard deviation across tasks.}
    \label{fig:perf-key}
\end{figure}

\subsection{Vision: Investigations on Keypoint Tokens}
\label{sec:exp-vision}
\textbf{What is the optimal number of keypoints to extract?}
In this work we proposed a pipeline, based on the DINO family of Vision Foundation Models, to transform a visual observation $o_i$ into a \textit{sequence of 3D keypoints $k_{1:K}$}. The number of extracted keypoints $K$ is an influential hyperparameters: a small $K$ reduces input dimensionality, but generates a strong information bottleneck and may fail to capture more nuanced geometric and semantic information from visual observations. On the other hand, a large $K$ allows us to capture many visual details, but it is then more susceptible to noise and visual distractors.
We measured the performance of KAT on a subset of the 9 tasks (\textit{Align T, Wipe Plate, Espresso, Pick and Place}) as a function of $K$, providing 10 demonstrations per task. We repeat the experimental procedure described in Sec. \ref{sec:few-shot-results} 4 times, and extract $K \in \{5,10,20,40\}$ keypoints from the observations, varying it at each experiment. We measure the performance on 10 test time episodes, and report mean and standard deviation in the plot of Fig. \ref{fig:perf-key}. The results show that the optimal $K$ lies between 10 and 20, with minimal variations in that range.

\begin{figure}[t]
    \centering
    \includegraphics[width=.5\textwidth]{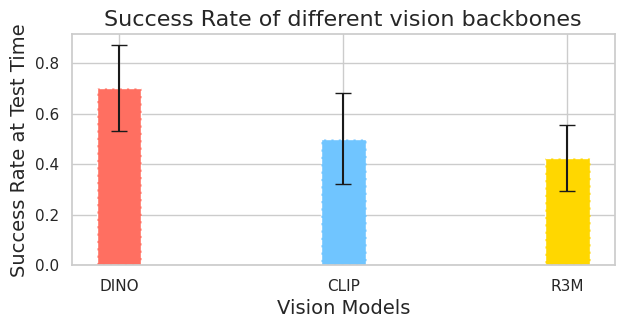}
    \caption{Success rate of our method when changing the vision backbone model from which we extract \textit{keypoint tokens}. Plot shows mean and standard deviation across tasks.}
    \label{fig:perf-vision-model}
\end{figure}

\textbf{What is the best vision model to extract descriptors from?}
To motivate our choice of using DINO-ViTs as vision backbones, we compared them to two other popular models in the robotics and computer vision literature: \textbf{CLIP} \cite{radford2021clip} and \textbf{R3M} \cite{nair2022r3m}. We apply the same algorithm described in Fig. \ref{fig:keypoint-tokens}, with the difference that the dense descriptor tensors are extracted from CLIP or R3M (in the Supplementary Material we provide more information on this process). These experiments allowed us to test the robustness and representation abilities of these hidden representations.
We run the same experimental setup described above in the previous subsection, setting $K = 10$ and training and testing on \textit{Align T, Wipe Plate, Espresso, Pick and Place}. Our results (Fig. \ref{fig:perf-vision-model}) demonstrate that DINO is the optimal choice, coherently with the results from the recent literature \cite{amir2022dinokp}. While R3M is pretrained on data more related to robotics (first person videos of humans performing manipulation tasks \cite{grauman2022ego4d}), interestingly it does not perform better than DINO, pretrained on a large task-agnostic dataset of web images \cite{caron2021dino}. This apparently counter-intuitive result was also reported in \cite{dasari2023unbiased}.

\begin{figure}[t]
    \centering
    \includegraphics[width=.5\textwidth]{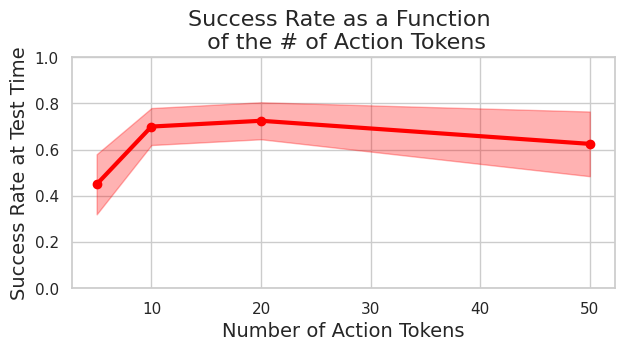}
    \caption{Success rate of our method as a function of the number of action tokens extracted from each demonstration trajectory, that strongly correlates with the number of action tokens generated at test time. Plot shows mean and standard deviation across tasks.}
    \label{fig:perf-act}
\end{figure}

\begin{figure}[t]
    \centering
    \includegraphics[width=.5\textwidth]{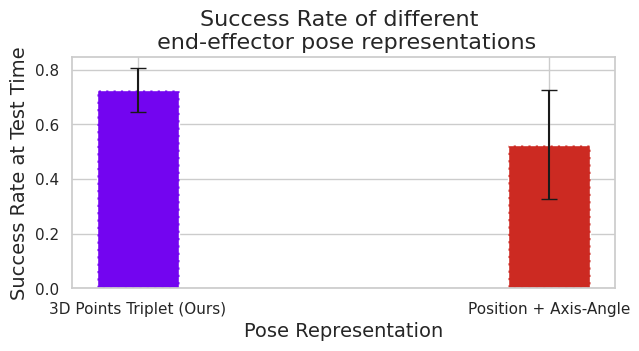}
    \caption{Success rate of our method with different representations of the end-effector poses. Plot shows mean and standard deviation across tasks.}
    \label{fig:actrepr-perf}
\end{figure}

\begin{figure}[t]
    \centering
    \includegraphics[width=.5\textwidth]{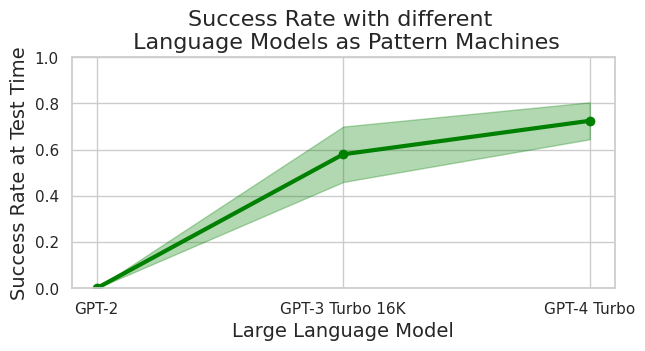}
    \caption{Success rate of KAT by varying the underlying LLM acting as the imitation learning machine. Plot shows mean and standard deviation across tasks.}
    \label{fig:llm-perf}
\end{figure}

\subsection{Action: Investigations on Action Tokens}
\textbf{What is the optimal number of actions to record and predict?}
In our work, actions are recorded as end-effector poses during demonstrations, then tokenised into \textit{action tokens}, each being a triplet of 3D positions in the world frame. Vice versa, at test time the Transformer predicts a series of \textit{action tokens} that are then transformed into end-effector poses and executed. An LLM Transformer is able to autonomously decide to stop the autoregressive generation of tokens, by generating an internal STOP token. The number of action tokens it predicts is often very close to the number of action tokens each demonstration was composed of.

Therefore, we investigate what is the optimal number of action tokens $N_a$ to provide for each demonstration, to guide the test time generation. A small $N_a$ reduces the length of the sequence and makes autoregressive generation easier and less prone to errors: however, it can also excessively subsample the trajectory of the end-effector, losing subtle but important movements. On the other hand, while a large $N_a$ allows to precisely capture movements, it also makes both the input sequences excessively long and difficult to process, and also the generation phase prone to errors, as common with sequence-to-sequence networks tasked to predict very long outputs.

We replicate the experimental setup of Sec. \ref{sec:exp-vision}. We record 10 demos on the \textit{Align T, Wipe Plate, Sweep, Bottle Upright} tasks at a very high frequency, and then subsample those to change the number of action tokens $N_a$ each demonstration is decomposed into, with $N_a \in {5,10,20,50}$. We then test the performance of KAT by providing to the Transformer these different quantities of tokenised actions extracted from the demonstrations, running 10 test time episodes per task.

Our results, in Fig. \ref{fig:perf-act}, show that the optimal number of tokens lies around 20: however, the drop at 50 is not large, suggesting the Transformer is still able to process this sequence length.

\textbf{What is the optimal representation of poses to use as action tokens?}
In Sec. \ref{sec:action-tokens} we introduced the way we tokenise end-effector poses, transforming them through an invertible mapping into a triplet of 3D points $\bm{e}^\mathcal{W}$. Is this the most effective representation to represent, process and generate sequences of actions? To motivate our design choice, we compared our tokenisation pipeline to the representation of poses as 3D position and axis-angle representation for orientation. This representation is therefore composed of 6 numbers instead of 9, but the orientation part lies in a different, non-Euclidean manifold. As described in Sec. \ref{sec:action-tokens}, all representations share an additional number $\delta_g$ representing the gripper open/closed state.

We replicate the above experimental setup, but compare the effect of representing and generating actions as action tokens $\bm{e}^\mathcal{W}$ or as position and axis-angle representation. Results in Figure \ref{fig:actrepr-perf} demonstrate that our choice outperforms the more classical position/axis-angle method of representing $SE(3)$ poses. As discussed in Sec. \ref{sec:action-tokens}, we hypothesise that representing poses as a set of Euclidean entities in the same frame as the keypoint tokens facilitates generalisation for learned imitation learning policies, that instead struggle more to generalise and interpolate/extrapolate correct axis-angle representations beyond the instances of the tasks observed during the demonstrations.

\subsection{Free Robotics Lunch: Better Imitation Learning Machines by Scaling Language Models}
Our work is inspired and designed around the finding that large Transformers, pretrained on sequence-to-sequence tasks, have the emergent capability to infer in few-shot, patterns belonging to different data modalities. This effect has been observed especially in Large Language Models. Scaling laws and empirical results have been obtained \cite{hoffmann2022training, brown2020language} that allow the predict the performance of these models as a function of the size of both their parameters and training dataset. 

As a result, the latest LLMs, like GPT-4 Turbo, can effectively act as few-shot imitation learning machines for robotics, reaching or surpassing the performance of state-of-the-art imitation techniques despite not needing any robotics data. As the field of LLMs has seen exponential growth in recent years \cite{brown2020language, touvron2023llama, openai2023gpt4, geminiteam2023gemini}, we hypothesise that future LLMs will be even better (as suggested by numbers reported for Gemini Ultra \cite{geminiteam2023gemini}, still unavailable at the time of writing), leading to even more efficient imitation learning machines. And this is without the need for any particular innovation in robotics itself, be it in terms of data or algorithms. To concretely support this hypothesis, we now evaluate and visualise the performance of different generations of LLMs.

We test KAT on the \textit{Align T, Wipe Plate, Sweep, Bottle Upright} tasks, selecting $K = 10$ and $N_a = 20$, but change the model of the LLM computing the output sequence given the demonstration sequences and the input sequence. We test three generations of \textbf{GPT} models: \textbf{GPT-2 \cite{gpt2}, GPT-3 Turbo 16k \cite{brown2020language}, GPT-4 Turbo \cite{openai2023gpt4}}. Results in Fig. \ref{fig:llm-perf} demonstrate that the evolution of these text-pretrained Transformers, albeit independent in any way from robotics, leads to an improvement in performance, as more recent models are more efficient at few-shot sequence-to-sequence pattern recognition and generation. This suggests that future models will be even more efficient, resulting in more effective \textit{general imitation learning machines}. This is a remarkable phenomenon, considering that it does not require gathering more robotics-tailored data, nor does it require actively researching fundamentally novel imitation learning algorithms.

A noteworthy aspect is the ability to effectively process longer sequences as inputs: GPT-2 was limited at 1024 word tokens, while later models saw a 10x-30x increase in a few years. Scaling the context input size of text-pretrained Transformers can help make in-context learning of trajectories more effective, tackling the bottleneck we observed in Fig. \ref{fig:num-demos}.

\section{Conclusion}
We introduced \textbf{Keypoint Action Tokens}, a framework that enables in-context imitation learning of human demonstrations, by repurposing large Transformers pretrained on text as general sequence-to-sequence learners. By tokenising both visual inputs and action outputs into a format suitable for text-based Transformers, we demonstrate that we can achieve state-of-the-art results in few-shot imitation learning on a series of challenging everyday tasks. We also analysed what design decisions result in the optimal performance of our method. Our work demonstrates that the progressive evolution of large Transformers pretrained on language, where data is abundant, can unexpectedly benefit the field of robot learning, where data is scarce.
\section{Limitations}

While achieving superior results in few-shot imitation learning with $\le 10-20$ demos, KAT does not currently scale as well as Diffusion Policies. We hypothesise that this is due to KAT relying entirely on in-context imitation learning, which can start acting suboptimally when the input sequences (the tokenised inputs and outputs of the demos, and the tokenised new observation) become excessively long. We believe that once $>50$ demos are available, it may be beneficial to finetune the pretrained Transformers instead of relying entirely on feed-forward in-context learning.

To tokenise the visual observations, we rely on extracting a predetermined amount of keypoints $K$. While we showed that this approach is effective, and even significantly improves the performance of end-to-end vision based Diffusion Policy, the method cannot adapt the number of extracted keypoints to different situations, where e.g. more objects are present. Future work into adaptable and dynamic keypoint extraction may improve the performance of keypoint-based methods, like KAT.

Given a prompt of length $\mathcal{L}$, the computational complexity of generating output tokens, based on the classic Transformer architecture \cite{vaswani2023attention}, is $\mathcal{O}(\mathcal{L}^2)$. Therefore, current Transformers will scale poorly to larger datasets of demonstrations in scenarios where online control is fundamental. However, there are several research avenues currently researching an efficient alternative to classic Transformers \cite{choromanski2022rethinking} or proposing alternatives which are equally effective but linear in complexity \cite{gu2023mamba}.

%% Use plainnat to work nicely with natbib. 

\bibliographystyle{plainnat}
\bibliography{main}

\begin{thebibliography}{40}
\providecommand{\natexlab}[1]{#1}
\providecommand{\url}[1]{\texttt{#1}}
\expandafter\ifx\csname urlstyle\endcsname\relax
  \providecommand{\doi}[1]{doi: #1}\else
  \providecommand{\doi}{doi: \begingroup \urlstyle{rm}\Url}\fi

\bibitem[Amir et~al.(2022)Amir, Gandelsman, Bagon, and Dekel]{amir2022dinokp}
Shir Amir, Yossi Gandelsman, Shai Bagon, and Tali Dekel.
\newblock Deep vit features as dense visual descriptors, 2022.
\newblock arXiv:2112.05814.

\bibitem[Caron et~al.(2021)Caron, Touvron, Misra, Jégou, Mairal, Bojanowski, and Joulin]{caron2021dino}
Mathilde Caron, Hugo Touvron, Ishan Misra, Hervé Jégou, Julien Mairal, Piotr Bojanowski, and Armand Joulin.
\newblock Emerging properties in self-supervised vision transformers, 2021.
\newblock arXiv:2104.14294.

\bibitem[Chi et~al.(2023)Chi, Feng, Du, Xu, Cousineau, Burchfiel, and Song]{chi2023diffusion}
Cheng Chi, Siyuan Feng, Yilun Du, Zhenjia Xu, Eric Cousineau, Benjamin Burchfiel, and Shuran Song.
\newblock Diffusion policy: Visuomotor policy learning via action diffusion, 2023.
\newblock arXiv:2303.04137.

\bibitem[Choromanski et~al.(2022)Choromanski, Likhosherstov, Dohan, Song, Gane, Sarlos, Hawkins, Davis, Mohiuddin, Kaiser, Belanger, Colwell, and Weller]{choromanski2022rethinking}
Krzysztof Choromanski, Valerii Likhosherstov, David Dohan, Xingyou Song, Andreea Gane, Tamas Sarlos, Peter Hawkins, Jared Davis, Afroz Mohiuddin, Lukasz Kaiser, David Belanger, Lucy Colwell, and Adrian Weller.
\newblock Rethinking attention with performers, 2022.
\newblock arXiv:2009.14794.

\bibitem[Clavera et~al.(2018)Clavera, Rothfuss, Schulman, Fujita, Asfour, and Abbeel]{clavera2018modelbased}
Ignasi Clavera, Jonas Rothfuss, John Schulman, Yasuhiro Fujita, Tamim Asfour, and Pieter Abbeel.
\newblock Model-based reinforcement learning via meta-policy optimization, 2018.
\newblock arXiv:1809.05214.

\bibitem[Collaboration(2023)]{embodimentcollaboration2023open}
Open X-Embodiment Collaboration.
\newblock Open x-embodiment: Robotic learning datasets and rt-x models, 2023.
\newblock arXiv:2310.08864.

\bibitem[Dasari et~al.(2023)Dasari, Srirama, Jain, and Gupta]{dasari2023unbiased}
Sudeep Dasari, Mohan~Kumar Srirama, Unnat Jain, and Abhinav Gupta.
\newblock An unbiased look at datasets for visuo-motor pre-training, 2023.
\newblock arXiv:2310.09289.

\bibitem[Doersch et~al.(2023)Doersch, Yang, Vecerik, Gokay, Gupta, Aytar, Carreira, and Zisserman]{doersch2023tapir}
Carl Doersch, Yi~Yang, Mel Vecerik, Dilara Gokay, Ankush Gupta, Yusuf Aytar, Joao Carreira, and Andrew Zisserman.
\newblock Tapir: Tracking any point with per-frame initialization and temporal refinement, 2023.

\bibitem[Dosovitskiy et~al.(2021)Dosovitskiy, Beyer, Kolesnikov, Weissenborn, Zhai, Unterthiner, Dehghani, Minderer, Heigold, Gelly, Uszkoreit, and Houlsby]{dosovitskiy2021vit}
Alexey Dosovitskiy, Lucas Beyer, Alexander Kolesnikov, Dirk Weissenborn, Xiaohua Zhai, Thomas Unterthiner, Mostafa Dehghani, Matthias Minderer, Georg Heigold, Sylvain Gelly, Jakob Uszkoreit, and Neil Houlsby.
\newblock An image is worth 16x16 words: Transformers for image recognition at scale, 2021.
\newblock arXiv:2010.11929.

\bibitem[Driess et~al.(2023)Driess, Xia, Sajjadi, Lynch, Chowdhery, Ichter, Wahid, Tompson, Vuong, Yu, Huang, Chebotar, Sermanet, Duckworth, Levine, Vanhoucke, Hausman, Toussaint, Greff, Zeng, Mordatch, and Florence]{driess2023palme}
Danny Driess, Fei Xia, Mehdi S.~M. Sajjadi, Corey Lynch, Aakanksha Chowdhery, Brian Ichter, Ayzaan Wahid, Jonathan Tompson, Quan Vuong, Tianhe Yu, Wenlong Huang, Yevgen Chebotar, Pierre Sermanet, Daniel Duckworth, Sergey Levine, Vincent Vanhoucke, Karol Hausman, Marc Toussaint, Klaus Greff, Andy Zeng, Igor Mordatch, and Pete Florence.
\newblock Palm-e: An embodied multimodal language model, 2023.
\newblock arXiv:2303.03378.

\bibitem[Duan et~al.(2017)Duan, Andrychowicz, Stadie, Ho, Schneider, Sutskever, Abbeel, and Zaremba]{duan2017oneshot}
Yan Duan, Marcin Andrychowicz, Bradly~C. Stadie, Jonathan Ho, Jonas Schneider, Ilya Sutskever, Pieter Abbeel, and Wojciech Zaremba.
\newblock One-shot imitation learning, 2017.
\newblock arXiv:1703.07326.

\bibitem[et~al.(2023{\natexlab{a}})]{brohan2023rt2}
Anthony~Brohan et~al.
\newblock Rt-2: Vision-language-action models transfer web knowledge to robotic control, 2023{\natexlab{a}}.
\newblock arXiv:2307.15818.

\bibitem[et~al.(2023{\natexlab{b}})]{geminiteam2023gemini}
Gemini~Team et~al.
\newblock Gemini: A family of highly capable multimodal models, 2023{\natexlab{b}}.
\newblock arXiv:2312.11805.

\bibitem[et~al.(2023{\natexlab{c}})]{touvron2023llama}
Hugo~Touvron et~al.
\newblock Llama 2: Open foundation and fine-tuned chat models, 2023{\natexlab{c}}.
\newblock arXiv:2307.09288.

\bibitem[et~al.(2022{\natexlab{a}})]{grauman2022ego4d}
Kristen~Grauman et~al.
\newblock Ego4d: Around the world in 3,000 hours of egocentric video, 2022{\natexlab{a}}.
\newblock arXiv:2110.07058.

\bibitem[et~al.(2022{\natexlab{b}})]{ahn2022saycan}
Michael~Ahn et~al.
\newblock Do as i can, not as i say: Grounding language in robotic affordances, 2022{\natexlab{b}}.
\newblock arXiv:2204.01691.

\bibitem[et~al.(2023{\natexlab{d}})]{openai2023gpt4}
OpenAI et~al.
\newblock Gpt-4 technical report, 2023{\natexlab{d}}.
\newblock arXiv:2303.08774.

\bibitem[et~al.(2022{\natexlab{c}})]{bommasani2022opportunities}
Rishi~Bommasani et~al.
\newblock On the opportunities and risks of foundation models, 2022{\natexlab{c}}.
\newblock arXiv:2108.07258.

\bibitem[et~al.(2020)]{brown2020language}
Tom B.~Brown et~al.
\newblock Language models are few-shot learners, 2020.
\newblock arXiv:2005.14165.

\bibitem[Goyal et~al.(2023)Goyal, Xu, Guo, Blukis, Chao, and Fox]{goyal2023rvt}
Ankit Goyal, Jie Xu, Yijie Guo, Valts Blukis, Yu-Wei Chao, and Dieter Fox.
\newblock Rvt: Robotic view transformer for 3d object manipulation, 2023.

\bibitem[Gu and Dao(2023)]{gu2023mamba}
Albert Gu and Tri Dao.
\newblock Mamba: Linear-time sequence modeling with selective state spaces, 2023.
\newblock arXiv:2312.00752.

\bibitem[Hoffmann et~al.(2022)Hoffmann, Borgeaud, Mensch, Buchatskaya, Cai, Rutherford, de~Las~Casas, Hendricks, Welbl, Clark, Hennigan, Noland, Millican, van~den Driessche, Damoc, Guy, Osindero, Simonyan, Elsen, Rae, Vinyals, and Sifre]{hoffmann2022training}
Jordan Hoffmann, Sebastian Borgeaud, Arthur Mensch, Elena Buchatskaya, Trevor Cai, Eliza Rutherford, Diego de~Las~Casas, Lisa~Anne Hendricks, Johannes Welbl, Aidan Clark, Tom Hennigan, Eric Noland, Katie Millican, George van~den Driessche, Bogdan Damoc, Aurelia Guy, Simon Osindero, Karen Simonyan, Erich Elsen, Jack~W. Rae, Oriol Vinyals, and Laurent Sifre.
\newblock Training compute-optimal large language models, 2022.
\newblock arXiv:2203.15556.

\bibitem[Huang et~al.(2022)Huang, Abbeel, Pathak, and Mordatch]{huang2022language}
Wenlong Huang, Pieter Abbeel, Deepak Pathak, and Igor Mordatch.
\newblock Language models as zero-shot planners: Extracting actionable knowledge for embodied agents, 2022.
\newblock arXiv:2201.07207.

\bibitem[Huang et~al.(2023)Huang, Wang, Zhang, Li, Wu, and Fei-Fei]{huang2023voxposer}
Wenlong Huang, Chen Wang, Ruohan Zhang, Yunzhu Li, Jiajun Wu, and Li~Fei-Fei.
\newblock Voxposer: Composable 3d value maps for robotic manipulation with language models, 2023.
\newblock arXiv:2307.05973.

\bibitem[James et~al.(2018)James, Bloesch, and Davison]{james2018task}
Stephen James, Michael Bloesch, and Andrew~J Davison.
\newblock Task-embedded control networks for few-shot imitation learning.
\newblock In \emph{Conference on robot learning}, pages 783--795. PMLR, 2018.

\bibitem[Khansari et~al.(2020)Khansari, Kappler, Luo, Bingham, and Kalakrishnan]{khansari2020action}
Mohi Khansari, Daniel Kappler, Jianlan Luo, Jeff Bingham, and Mrinal Kalakrishnan.
\newblock Action image representation: Learning scalable deep grasping policies with zero real world data, 2020.
\newblock arXiv:2005.06594.

\bibitem[Kwon et~al.(2023)Kwon, Palo, and Johns]{kwon2023language}
Teyun Kwon, Norman~Di Palo, and Edward Johns.
\newblock Language models as zero-shot trajectory generators, 2023.
\newblock arXiv:2310.11604.

\bibitem[Liang et~al.(2023)Liang, Huang, Xia, Xu, Hausman, Ichter, Florence, and Zeng]{liang2023code}
Jacky Liang, Wenlong Huang, Fei Xia, Peng Xu, Karol Hausman, Brian Ichter, Pete Florence, and Andy Zeng.
\newblock Code as policies: Language model programs for embodied control, 2023.
\newblock arXiv:2209.07753.

\bibitem[Mirchandani et~al.(2023)Mirchandani, Xia, Florence, Ichter, Driess, Arenas, Rao, Sadigh, and Zeng]{mirchandani2023large}
Suvir Mirchandani, Fei Xia, Pete Florence, Brian Ichter, Danny Driess, Montserrat~Gonzalez Arenas, Kanishka Rao, Dorsa Sadigh, and Andy Zeng.
\newblock Large language models as general pattern machines, 2023.
\newblock arXiv:2307.04721.

\bibitem[Nair et~al.(2022)Nair, Rajeswaran, Kumar, Finn, and Gupta]{nair2022r3m}
Suraj Nair, Aravind Rajeswaran, Vikash Kumar, Chelsea Finn, and Abhinav Gupta.
\newblock R3m: A universal visual representation for robot manipulation, 2022.
\newblock arXiv:2203.12601.

\bibitem[Olsson et~al.(2022)Olsson, Elhage, Nanda, Joseph, DasSarma, Henighan, Mann, Askell, Bai, Chen, et~al.]{olsson2022context}
Catherine Olsson, Nelson Elhage, Neel Nanda, Nicholas Joseph, Nova DasSarma, Tom Henighan, Ben Mann, Amanda Askell, Yuntao Bai, Anna Chen, et~al.
\newblock In-context learning and induction heads.
\newblock \emph{arXiv prnote arXiv:2209.11895}, 2022.

\bibitem[Radford et~al.(2019)Radford, Wu, Child, Luan, Amodei, and Sutskever]{gpt2}
Alec Radford, Jeffrey Wu, Rewon Child, David Luan, Dario Amodei, and Ilya Sutskever.
\newblock Language models are unsupervised multitask learners, 2019.

\bibitem[Radford et~al.(2021)Radford, Kim, Hallacy, Ramesh, Goh, Agarwal, Sastry, Askell, Mishkin, Clark, Krueger, and Sutskever]{radford2021clip}
Alec Radford, Jong~Wook Kim, Chris Hallacy, Aditya Ramesh, Gabriel Goh, Sandhini Agarwal, Girish Sastry, Amanda Askell, Pamela Mishkin, Jack Clark, Gretchen Krueger, and Ilya Sutskever.
\newblock Learning transferable visual models from natural language supervision, 2021.
\newblock arXiv:2103.00020.

\bibitem[Vaswani et~al.(2023)Vaswani, Shazeer, Parmar, Uszkoreit, Jones, Gomez, Kaiser, and Polosukhin]{vaswani2023attention}
Ashish Vaswani, Noam Shazeer, Niki Parmar, Jakob Uszkoreit, Llion Jones, Aidan~N. Gomez, Lukasz Kaiser, and Illia Polosukhin.
\newblock Attention is all you need, 2023.
\newblock arXiv:1706.03762.

\bibitem[Vecerik et~al.(2023)Vecerik, Doersch, Yang, Davchev, Aytar, Zhou, Hadsell, Agapito, and Scholz]{vecerik2023robotap}
Mel Vecerik, Carl Doersch, Yi~Yang, Todor Davchev, Yusuf Aytar, Guangyao Zhou, Raia Hadsell, Lourdes Agapito, and Jon Scholz.
\newblock Robotap: Tracking arbitrary points for few-shot visual imitation, 2023.

\bibitem[Vemprala et~al.(2023)Vemprala, Bonatti, Bucker, and Kapoor]{vemprala2023chatgpt}
Sai Vemprala, Rogerio Bonatti, Arthur Bucker, and Ashish Kapoor.
\newblock Chatgpt for robotics: Design principles and model abilities, 2023.
\newblock arXiv:2306.17582.

\bibitem[von Oswald et~al.(2023)von Oswald, Niklasson, Randazzo, Sacramento, Mordvintsev, Zhmoginov, and Vladymyrov]{vonoswald2023transformers}
Johannes von Oswald, Eyvind Niklasson, Ettore Randazzo, João Sacramento, Alexander Mordvintsev, Andrey Zhmoginov, and Max Vladymyrov.
\newblock Transformers learn in-context by gradient descent, 2023.
\newblock arXiv:2212.07677.

\bibitem[Vosylius and Johns(2023)]{vosylius2023fewshot}
Vitalis Vosylius and Edward Johns.
\newblock Few-shot in-context imitation learning via implicit graph alignment, 2023.
\newblock arXiv:2310.12238.

\bibitem[Yu et~al.(2023)Yu, Gileadi, Fu, Kirmani, Lee, Arenas, Chiang, Erez, Hasenclever, Humplik, Ichter, Xiao, Xu, Zeng, Zhang, Heess, Sadigh, Tan, Tassa, and Xia]{yu2023language}
Wenhao Yu, Nimrod Gileadi, Chuyuan Fu, Sean Kirmani, Kuang-Huei Lee, Montse~Gonzalez Arenas, Hao-Tien~Lewis Chiang, Tom Erez, Leonard Hasenclever, Jan Humplik, Brian Ichter, Ted Xiao, Peng Xu, Andy Zeng, Tingnan Zhang, Nicolas Heess, Dorsa Sadigh, Jie Tan, Yuval Tassa, and Fei Xia.
\newblock Language to rewards for robotic skill synthesis, 2023.
\newblock arXiv:2306.08647.

\bibitem[Zhou et~al.(2023)Zhou, Dean, Srirama, Rajeswaran, Pari, Hatch, Jain, Yu, Abbeel, Pinto, Finn, and Gupta]{zhou2023toto}
Gaoyue Zhou, Victoria Dean, Mohan~Kumar Srirama, Aravind Rajeswaran, Jyothish Pari, Kyle Hatch, Aryan Jain, Tianhe Yu, Pieter Abbeel, Lerrel Pinto, Chelsea Finn, and Abhinav Gupta.
\newblock Train offline, test online: A real robot learning benchmark, 2023.
\newblock arXiv:2306.00942.

\end{thebibliography}

\newpage
.
\newpage
\section{Appendix}

\subsection{Tasks Success Criteria}
Here we describe the adopted criteria to define an episode as successful for each task.

\begin{itemize}
    \item \textbf{Align T}: Given an imaginary frame of reference with axis $x$ and $y$ aligned with the longer and shorter sides of the table respectively, the long part of the T is withing 20 degrees to the $y$ side.
    
    \item \textbf{Wiping a Plate}: The sponge grasped by the robot touched at least $80\%$ of the edge of the plate.
    
    \item \textbf{Sweep}: The object to be swept is entirely on the dustpan at the end of the episode.
    
    \item \textbf{Espresso}: The espresso capsule is inserted in the slot and the lid is closed.
    
    \item \textbf{Scooping}: The robot scoops at least 5 grams of chocolate powder from the cup.
    
    \item \textbf{Pick and Pour}: The French press is inclined of at least $50$ degrees from its normal vertical position while being held by the robot, with the spout pointing inside the mug. For practical reasons the French press was not filled with liquid, but we mimic a movement that would pour its contents into the mug.

    \item \textbf{Hang}: The cloth hanger its stably placed on its support.
    
    \item \textbf{Put Bottle Upright}: At the end of the episode, the bottle is in a stable vertical position when no longer grasped.
    
    \item \textbf{Pick and Place}: The apple (or orange) is entirely inside the red container.
\end{itemize}

\subsection{Collecting Demos and Train-Test Distributions}
When collecting demos,  we randomise the pose of the objects on the table. In particular, we move the objects on a $50 cm \times 30 cm$ area, and randomise their orientation around the imaginary vertical axis in a $45$ degrees range from a "standard orientation". To test for extrapolation at test time, we move objects in a larger area of $70 cm \times 40 cm$ and rotate them in a $60$ degrees range.

For tasks that can be tested for objects generalisation (\textit{Wiping a Plate, Sweep, Scooping, Pick and Pour, Bottle Upright, Pick and Place} we provide demonstrations using 2 different sets of objects, and then test on both these sets and a third unseen set.

When testing for robustness to distractors, we change our background, which is normally a black cloth, to expose the wooden table underneath, and position an additional set of 2-3 random objects of different classes then the ones involved in the task, randomly positioned as the other objects.

\subsection{Prompting the Language Model}

We mainly use GPT-4 Turbo for our experiments, except when explicitly comparing it to GPT-3 and GPT-2. For GPT-4 and GPT-3 we use the official OpenAI API. For GPT-2, we use the Transformers library by HuggingFace to compute its outputs locally.

While largely receiving tokenised numbers as input, the way GPT-4 is structured requires a "\textit{system}" message describing what its task is. We therefore briefly describe that GPT-4 is tasked to solve a time series pattern recognition task, receiving examples of sequence-to-sequence data, and needing to find an output sequence given an input sequence that emulates the underlying patterns of the inputs. Notice we \textit{do not mention robotics or tasks anywhere}: the model is only asked to act as a sequence-to-sequence general pattern recognition machine.

In particular, the instruction is 

\textit{"You are a pattern generator machine. I will give you a series of patterns with INPUTS and OUTPUTS as examples. Then, you will receive a new INPUTS, and you have to generate OUTPUTS following the pattern that appears in the data. 
The points are (x,y,z) coordinates.
Only reply with the OUTPUTS numbers."
}

When comparing Language Models, we use the same prompt for each, composed of the above, plus the tokenised demonstrations as sequences of keypoints and action tokens, and finally the test time sequence of new keypoint tokens.

%\subsection{Implementation of Keypoint and Action Tokens}
%We will provide an implementation of both the keypoint and action tokenisation pipelines on our website \href{https://sites.google.com/view/keypoint-action-tokens}{https://sites.google.com/view/keypoint-action-tokens}. While it relies on some specific robot and setup dependent information, like camera calibration, we provide an example implementation abstracting away these details that can then be implemented by the user.

\subsection{Extracting descriptors from DINO, CLIP and R3M}
Here we describe in more detail how we extract the dense descriptor maps, given an input observation, to then extract keypoints.
For DINO, we use \textit{DINO-ViTs8}, with a stride of 4. We extract the \textit{key} descriptors from the 9th layer.
For CLIP, we use the CLIP RN50 version, and extract the output of the fifth-to-last layer of the ResNet50 backbone, a 3 dimensional tensor.
Likewise for R3M, we extract the output of the fifth-to-last layer of the ResNet50 backbone, a 3 dimensional tensor.
We explored each of these hyperparameters to maximise the performance of each of these methods.

\subsection{Implementation of Diffusion Policy}
As mentioned in the main paper, we rely on the official implementation by the authors of Diffusion Policy to train it as a baseline.
%\href{https://colab.research.google.com/drive/1gxdkgRVfM55zihY9TFLja97cSVZOZq2B?usp=sharing}{https://colab.research.google.com/drive/
%1gxdkgRVfM55zihY9TFLja97cSVZOZq2B?usp=sharing}.

We adapt the input and output spaces to our tasks, using vision for the end-to-end baseline and the state-based for the KeyAct baseline that takes keypoints as input. We modify mostly the training parameters, such as batch size and epochs, to maximise performance by evaluating both real world results and train/validation losses.

\end{document}